\documentclass[runningheads]{llncs}
\usepackage[T1]{fontenc}
\usepackage{graphicx}
\usepackage{hyperref}
\usepackage{tabularx}
\usepackage{makecell}
\usepackage{booktabs}
\usepackage{amsmath}
\usepackage{booktabs}
\usepackage{graphicx}
\usepackage{xspace}
\usepackage{amsmath}
\usepackage{amssymb}
\usepackage{pifont}
\newcommand{\sys}{RogueRover\xspace}
\usepackage{microtype}

\begin{document}

\title{RogueRover: Autonomous Rogue Device Localization for Incident Response}

\author{Priyanka Prakash Surve\inst{1}\orcidID{0009-0001-5673-2687} \and
Asaf Shabtai\inst{1}\orcidID{0000-0003-0630-4059} \and
Yuval Elovici\inst{1}\orcidID{0000-0002-9641-128X}}

\authorrunning{Surve et al.}

\institute{Ben Gurion University of the Negev\\
\email{surve@post.bgu.ac.il}\\
\email{shabtaia@bgu.ac.il}
\email{elovici@bgu.ac.il}
}

\maketitle              

\begin{abstract}
Physically localizing unauthorized wireless devices remains a critical bottleneck in cyber-physical security operations, where rogue access points can provide entry points for lateral movement and persistent compromise. While such devices can often be detected through network-side mechanisms, determining their physical location typically requires dense sensing infrastructure, site-specific RF fingerprinting, or manual inspection, limiting timely incident response.

We investigate whether a single commodity robot can autonomously detect and localize rogue wireless devices under zero-configuration constraints, without RF fingerprinting, pre-installed sensors, or site calibration.

We present \sys{},\footnote{Data, code, and a demonstration video are
available at \url{https://github.com/PriyankaPSurve/rogue-rover.} } an end-to-end system in which a quadruped robot autonomously patrols, collects spatially labeled RSSI measurements via a standard 802.11 interface, and estimates device locations offline. We evaluate the system across 11 patrol runs in a real indoor environment, with 6 rogue devices deployed under heterogeneous propagation conditions. Across 62 AP–patrol sessions, \sys{} achieves a median single-patrol localization error of 1.62 m without prior RF knowledge. Under multi-run aggregation, five of six devices are localized within 1 m. A blind trial validates the full pipeline, correctly identifying rogue devices among 73 observed BSSIDs and localizing them with errors of 0.34 m and 1.84 m.

Across environments, simple weighted-centroid estimators perform comparably to, or better than, parametric path-loss models, indicating that measurement coverage from autonomous patrols is the primary determinant of localization accuracy under zero-prior constraints.

Our results demonstrate that infrastructure-free, autonomous localization is feasible in practice, enabling rapid physical incident response in cyber-physical environments without additional sensing infrastructure. All measurement data and analysis code are released to support reproducibility.

\keywords{Cyber-Physical Systems Security  \and Incident Response \and Wireless Security.}
\end{abstract}
\section{Introduction}
\label{sec:introduction}

Industrial and enterprise cyber-physical systems (CPS) increasingly rely on networked wireless infrastructure for monitoring, control, and coordination. 
In such environments, unauthorized wireless devices, including rogue access points (RAPs), introduce persistent attack surfaces by enabling covert network access, lateral movement, and policy bypass. 
While these devices can often be detected through network-side mechanisms such as whitelist comparison, a critical bottleneck in incident response remains their \emph{physical localization}, namely determining the device’s location within a facility so that it can be investigated and removed.

In practice, physically locating rogue devices is challenging under realistic deployment constraints.
Static WIDS/WIPS systems require dense sensing infrastructure and may leave coverage gaps~\cite{meraki_air_marshal}. 
Fingerprinting-based methods achieve high accuracy but require site-specific calibration and maintenance as environments evolve~\cite{bahl2000radar,youssef2005horus,jang2018indoor}. 
CSI-based techniques offer sub-meter accuracy but depend on specialized hardware and firmware, while multi-robot approaches improve coverage at the cost of coordination complexity~\cite{yang2013rssi,awad2018collaborative}. 
These limitations hinder rapid and infrastructure-independent incident response in dynamic CPS environments.

We examine whether a single-commodity mobile robot can address these constraints and enable autonomous physical localization without prior configuration.
We present \sys{}, an end-to-end system in which a quadruped robot autonomously patrols a pre-mapped facility, collects spatially labeled RSSI measurements via a standard 802.11 interface, and estimates the locations of unauthorized wireless devices via an offline analysis pipeline.
By decoupling localization from pre-installed infrastructure and calibration, \sys{} enables deployment as an on-demand incident response tool.

We evaluate \sys{} across eleven patrol runs in an 800\,m$^2$ indoor environment with six rogue APs deployed across open, glass-partitioned, and drywall-enclosed areas.
A single patrol achieves a median localization error of 1.62\,m without prior RF knowledge.
A blind evaluation on two previously unseen devices confirms generalization beyond the primary dataset.
Across 62 AP — patrol sessions, simple weighted-centroid estimators~\cite{blumenthal2007weighted} perform comparably to parametric regression models~\cite{zhuang2015wireless}, indicating that spatial measurement coverage obtained through autonomous patrol is the dominant factor influencing localization accuracy.

\noindent\textbf{Contributions.}

\begin{enumerate}

\item \textbf{Infrastructure-free localization system.}
We present an autonomous system that localizes rogue wireless devices without RF fingerprinting, pre-installed sensors, or site calibration.

\item \textbf{Zero-configuration feasibility.}
We demonstrate that a single-commodity robot achieves a median localization error of 1.62\,m during a single patrol.

\item \textbf{Bridging detection and physical remediation.}
We show that autonomous patrol enables practical localization of detected rogue devices, closing the gap between network-level detection and physical incident response.

\end{enumerate}

Existing detection mechanisms identify unauthorized devices but do not provide their physical location, leaving a gap between network-level detection and physical remediation. \sys{} addresses this gap by enabling autonomous, infrastructure-free localization suitable for incident response workflows.
\section{Related Work}
The problem of rogue AP mitigation spans three decades, with approaches differentiated primarily by their deployment assumptions.

\emph{Infrastructure-based detection.}
Commercial WIDS/WIPS systems such as Cisco Air Marshal~\cite{meraki_air_marshal} deploy scanning radios co-located with managed APs and detect rogue SSIDs via wired-side MAC correlation.
Ma~et~al.~\cite{ma2008hybrid} combine wireless monitoring with wired-side traffic fingerprinting for rogue detection without specialized hardware.
Arisandi~et~al.~\cite{arisandi2025invisible} show that Layer-2 frame anomalies provide complementary signals on commodity interfaces.
These systems focus on detection and do not directly provide physical localization.

\emph{Fingerprinting and CSI-based localization.}
RADAR~\cite{bahl2000radar} and Horus~\cite{youssef2005horus} build site-specific radio maps to achieve sub-2m accuracy, but environmental changes require repeated calibration~\cite{jang2018indoor}.
CSI-based methods~\cite{yang2013rssi,wu2012csi} achieve sub-meter accuracy but require modified firmware on specific network interfaces.
Both approaches assume that infrastructure or calibration has been performed prior to rogue deployment, which may not be available during incident response.

\emph{RSSI-based AP localization.}
Koo and Cha~\cite{koo2010localizing} approximate the RSSI-distance relationship and apply multilateration, requiring at least four measurement points; their evaluation is simulation-based.
Zhuang~et~al.~\cite{zhuang2015wireless} jointly estimate AP position and path-loss parameters via nonlinear least squares without prior propagation knowledge, but report reduced accuracy near room boundaries~\cite{awad2018collaborative}.
Our LS+Huber baseline employs a comparable zero-prior formulation and shows reduced accuracy under sparse, noisy measurements.

\emph{Mobile robots for WiFi sensing.}
Zhang~et~al.~\cite{zhang2020wifi} localize a robot using known infrastructure APs with offline RSS fingerprinting, addressing a different problem formulation.
Parashar and Parasuraman~\cite{parashar2020particle} localize a WiFi AP via direction-of-arrival estimation on a mobile robot, requiring directional hardware beyond standard omnidirectional 802.11n interfaces.
Awad~et~al.~\cite{awad2018collaborative} demonstrate AP localization using a four-robot swarm in 10$\times$10\,m spaces, reporting sub-meter errors under controlled placement. Their prior single-robot study~\cite{awad2016wimap} reports similar accuracy under controlled conditions.
Performance in larger environments depends strongly on measurement geometry and AP placement.

RogueRover addresses the problem of localizing an unknown rogue emitter using a single commodity robot without prior RF calibration, offline training, or specialized hardware.
\section{System Design}
\label{sec:system_design}

\subsection{Threat Model}
\label{subsec:threat_model}

\textbf{Defender.}
The organizational security team operates within a managed cyber-physical environment (e.g., enterprise, industrial, or restricted facility) and possesses:
(i) an occupancy grid map of the facility,
(ii) network management records identifying authorized infrastructure, and
(iii) a commodity mobile robot with LiDAR-based navigation and dual 802.11n interfaces, one for passive scanning and one for robot communications.
The defender has no pre-installed wireless sensors, no RF fingerprint database, no specialized antennas, and no prior knowledge of RF propagation characteristics.
The goal of the defender is to enable rapid \emph{physical incident response}, namely, locating and removing unauthorized wireless devices once detected.

\noindent\textbf{Adversary.}
We consider an adversary that deploys a rogue wireless access point within the facility to establish an unauthorized communication channel.
Such devices may be used to bypass network controls, enable lateral movement, or provide persistent external access to internal systems.
We assume the rogue AP operates using standard 802.11 behavior and emits beacon frames required for client association.
The adversary may choose arbitrary placement and transmit power, but does not actively coordinate with the robot or adapt in real time to its presence.

We focus on scenarios in which rogue devices remain active for operationally relevant durations (e.g., hours to days), reflecting conditions such as shadow IT deployments or unattended attacker infrastructure.
This captures cases where detection occurs first, and physical localization is required for remediation.

\noindent\textbf{Out of scope.}
We do not consider adversaries that dynamically disable transmission in response to patrol, frequently change identity (e.g., rapid MAC rotation), or actively evade localization through coordinated behavior.
We also limit our evaluation to single-floor environments and stationary devices operating on the 2.4\,GHz band.
These constraints allow us to isolate the feasibility of infrastructure-free localization under realistic but controlled conditions.

\subsection{Architecture}
\label{subsec:architecture}

\begin{figure*}[t]
  \centering
  \includegraphics[width=\textwidth]{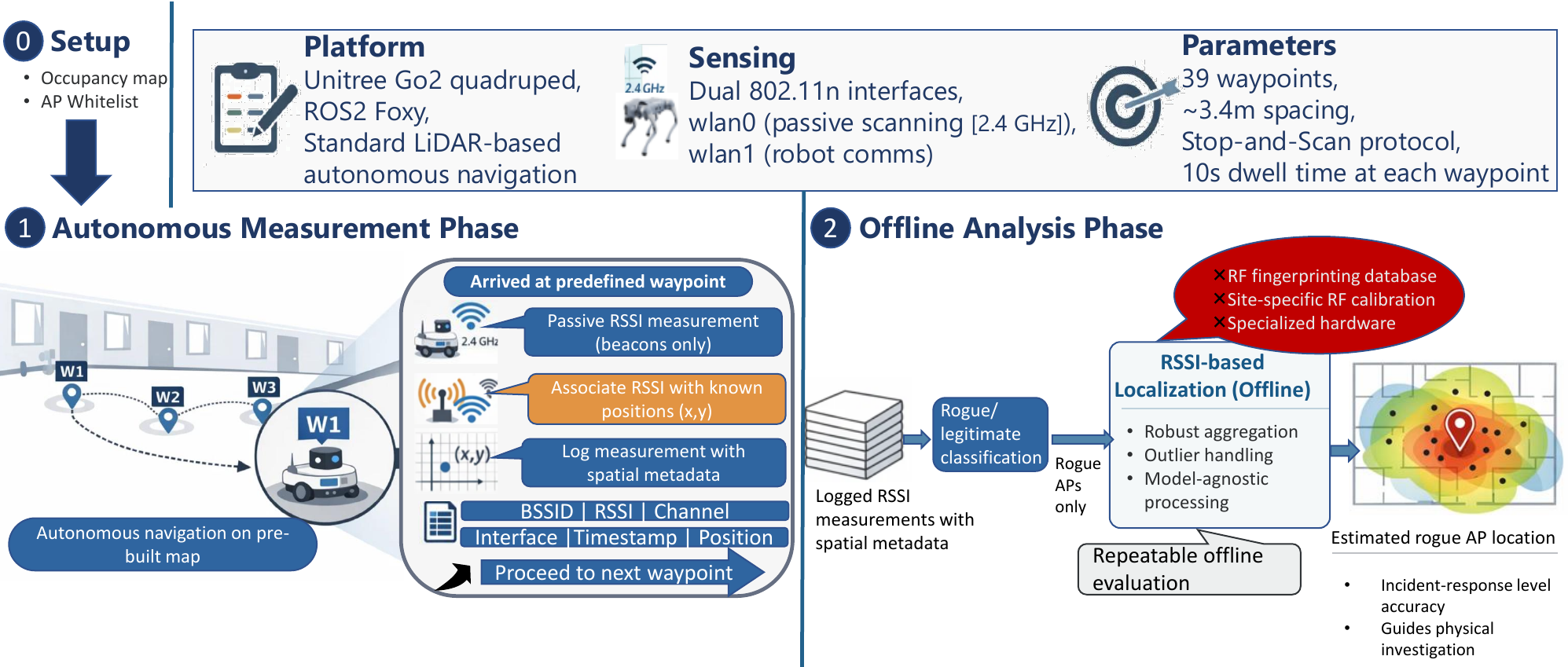}
  \caption{Spatially labeled RSSI measurements collected at patrol waypoints (39 in our evaluation). The offline phase performs rogue detection followed by RSSI-based localization without RF fingerprinting, pre-installed sensors, or site calibration.}
  \label{fig:arch}
\end{figure*}

As shown in Figure~\ref{fig:arch}, \sys{} separates navigation from signal collection.
The robot follows a predefined sequence of $n$ waypoints derived from the occupancy map and halts at each location to perform a passive 802.11 beacon scan (Stop-and-Scan protocol).
Halting ensures that each RSSI measurement is associated with a fixed coordinate under stable signal conditions during the 10\,second dwell period.

For each detected BSSID~$b$, the patrol produces a spatially labeled measurement set:
\begin{equation}
  \mathcal{D}_b = \{(x_i,\, y_i,\, r_i)\}_{i=1}^{N_b}
\end{equation}
where $(x_i, y_i)$ denotes the robot's localized position at waypoint~$i$, and $r_i \in \mathbb{R}$ is the observed RSSI in dBm.
Measurements are logged with BSSID, channel, interface, and timestamp and processed offline.

This separation enables \sys{} to operate as an on-demand incident response tool: the robot performs a single patrol to collect measurements, after which detection and localization are executed without requiring persistent infrastructure.

\subsection{Detection}
\label{subsec:detection}

Given the defender’s objective of identifying unauthorized devices for physical remediation, detection prioritizes low-overhead heuristics that require no site-specific training.

For each observed BSSID~$b$, we compute a heuristic rogue score using three lightweight signals:

\noindent \emph{(1) Locally administered MAC address.}
If the IEEE locally administered bit is set, the address is assigned or randomized by software.
This behavior is common for consumer hotspot devices and less typical for managed infrastructure.
A locally administered address contributes $+1.5$.

\noindent \emph{(2) SSID patterns.}
SSIDs matching common mobile hotspot patterns (e.g., \texttt{Galaxy\_*}, \texttt{AndroidAP*}) contribute $+1.0$, while SSIDs matching known infrastructure names in whitelist~$\mathcal{W}$ contribute $-1.0$.

\noindent \emph{(3) Vendor lookup.}
A MAC vendor lookup assigns $-0.5$ to manufacturers associated with managed infrastructure.

The total score is the sum of these contributions.
BSSIDs with score $\geq 0.5$ are flagged as potential rogue candidates.
Score weights are fixed \emph{a priori} and not tuned on evaluation data.

To ensure sufficient spatial diversity for localization, flagged devices must appear at least $k_{\min}=3$ distinct waypoints.
Devices below this threshold are reported as detected but not localizable, reflecting insufficient spatial evidence for reliable estimation.

\emph{Limitation.}
The scoring heuristic does not address BSSID spoofing or adaptive adversaries.
An attacker cloning a whitelisted MAC address or dynamically altering identifiers may evade detection.
Addressing such behavior (e.g., via clock-skew analysis or cryptographic validation) is orthogonal to the localization problem and left for future work.

\subsection{Localization Method Selection}
\label{subsec:method_selection}

Given $\mathcal{D}_b$, we estimate the AP position $(\hat{x}, \hat{y})$ without prior knowledge of transmit power or path-loss parameters.
This zero-prior constraint reflects the incident response setting, where rogue devices appear without warning and no calibration data is available.

We structure our method selection around three questions.

\noindent\textbf{Why not fingerprinting?}
Fingerprint-based methods require a pre-collected RSSI map of the facility at known positions.
Maintaining such maps incurs a high operational cost and must be repeated as environments change.
In incident response scenarios, no prior radio map may exist.
We therefore exclude fingerprinting under the zero-prior constraint.

\noindent\textbf{Why not trilateration?}
Trilateration converts RSSI to distance using a path-loss model, requiring knowledge of transmit power and propagation parameters that are unknown for rogue devices.
Joint estimation is ill-conditioned in the presence of multipath and limited sampling.
In our evaluation, LS+Huber achieves a median error of 3.41\,m under zero-prior conditions, exceeding centroid-based methods.
We therefore treat parametric models as baselines.

\noindent\textbf{Why centroid methods?}
Centroid approaches require no parametric assumptions and rely only on the relative ordering of signal strength across spatial samples.
With sufficient spatial coverage, this assumption remains robust under obstruction and multipath, aligning naturally with the zero-prior constraint.

We evaluate three centroid variants—Weighted Centroid (WCL), Exponentially Weighted Centroid (ExpWCL), and an Ensemble method—alongside a parametric baseline.
Details are provided in~\S\ref{sec:localization}.
\section{Localization}
\label{sec:localization}

For each rogue AP with measurement set $\mathcal{D}_b = \{(x_i, y_i, r_i)\}_{i=1}^{N_b}$, where $(x_i, y_i)$ is the robot's position at waypoint~$i$ and $r_i$ is the RSSI in dBm, we estimate the AP position $(\hat{x}, \hat{y})$ using only passively collected signal strength measurements.

All methods are evaluated under a \emph{zero-prior constraint} consistent with the incident response setting: no knowledge of transmit power, no calibrated path-loss parameters, and no pre-existing RF map. This constraint directly influences method selection, excluding approaches that rely on prior calibration or specialized hardware.

\noindent\textbf{Pre-processing.}
We discard patrol points whose RSSI is more than 25\,dB below the session maximum.
This filter removes weak far-field measurements that introduce high variance under multipath conditions, while retaining at least three spatially distinct points for stable estimation.
When multiple patrols are available, we retain the maximum RSSI observed at each waypoint after session filtering to reduce temporal variability and transient attenuation effects.

\noindent\textbf{Weighted centroid (WCL).}
The AP position is estimated as the power-weighted average of waypoint positions:
\[
  \hat{x} = \frac{\sum_i w_i x_i}{\sum_i w_i}, \qquad
  \hat{y} = \frac{\sum_i w_i y_i}{\sum_i w_i},
\]
with weights $w_i = 10^{r_i/10}$ (linear mW).
WCL requires no propagation model or calibration and serves as a robust baseline under the zero-prior constraint.

\noindent\textbf{Exponential weighted centroid (ExpWCL).}
ExpWCL replaces the linear-power weights with $w_i = e^{\beta r_i}$, using $\beta = 0.5$.
This increases the influence of high-RSSI measurements: a 10\,dB difference yields a weight ratio of $e^5 \approx 148{:}1$, compared to $10{:}1$ in WCL.
This behavior emphasizes spatial regions with strong signal dominance, which is beneficial when patrol coverage includes points near the emitter.
ExpWCL remains parameter-light and requires no environment-specific calibration.

\noindent\textbf{Path-loss least squares (LS+Huber).}
For comparison, we include a parametric baseline based on the log-distance path-loss model $r(d)=P_0-10n\log_{10}(d)$.
The method jointly estimates the reference power $P_0$, the path-loss exponent $n$, and the AP position $(\hat{x}, \hat{y})$ using Huber-loss minimization with multi-start initialization.

In the absence of prior calibration, this problem is ill-conditioned due to the coupling between unknown parameters and spatial position, particularly under multipath propagation.
We constrain the optimization to a $\pm$10\,m region around the WCL estimate and exclude distant points to improve stability.
LS+Huber is evaluated under the same zero-prior constraint and serves to quantify the benefit (or lack thereof) of parametric modeling without prior information.

\noindent\textbf{Ensemble consensus.}
We combine the three estimates to improve robustness.
If two estimates differ by at most 5\,m, we average them and discard the third.
If all three differ, we compute their geometric median.
This approach mitigates individual estimator failures without requiring environment-specific tuning or parameter selection.

\noindent\textbf{Design rationale.}
The objective of this comparison is not to identify a globally optimal localization algorithm, but to evaluate whether increased model complexity improves accuracy under zero-prior, infrastructure-free conditions.
As shown in Section~\ref{sec:eval}, localization accuracy is dominated by spatial measurement coverage obtained through patrol, rather than estimator sophistication.
\section{Evaluation}
\label{sec:eval}

\subsection{Experimental Setup}
\label{subsec:setup}

We evaluate \sys{} across eleven autonomous patrol runs conducted over four days in a single-floor office environment of approximately 800\,m$^2$.
Each patrol follows a fixed 39-waypoint route derived from the SLAM occupancy map, with waypoints spaced approximately 3.4\,m apart.
At each waypoint, the robot halts for 10\,s and performs a passive \texttt{iw scan}, recording all visible 2.4\,GHz BSSIDs.
Each patrol lasts approximately 20\,minutes.

The environment includes background infrastructure APs and external signals from adjacent areas, all of which are observed during scanning and processed by the detection pipeline.

Six rogue APs are deployed across three propagation environments: open-space corridors (Galaxy~F62e1f3, Laptop-123456), drywall-enclosed rooms (Galaxy~hotspot, unitree-laptop), and glass-partition offices (Galaxy\_1395, Laptop-181938).
Devices include smartphones and laptops configured as mobile hotspots.
Ground-truth positions are measured using a laser distance meter.
All APs remain stationary during evaluation.

Across eleven patrols with six APs, 62 AP–session instances provide sufficient spatial observations for localization.
In four instances, the corresponding AP did not emit detectable beacon frames during that patrol, consistent with normal commodity hotspot behavior.
Localization requires at least $k_{\min}=3$ spatially distinct observations per session.
All available observations satisfying that requirement are included in the experiment.
Localization error is defined as the Euclidean distance between the estimated and ground-truth AP positions in meters.
We report median error and the proportion of sessions achieving sub-1\,m accuracy.

\subsection{Single-Patrol Accuracy}
\label{subsec:single}

\begin{figure}[t]
  \centering
  \includegraphics[width=\columnwidth]{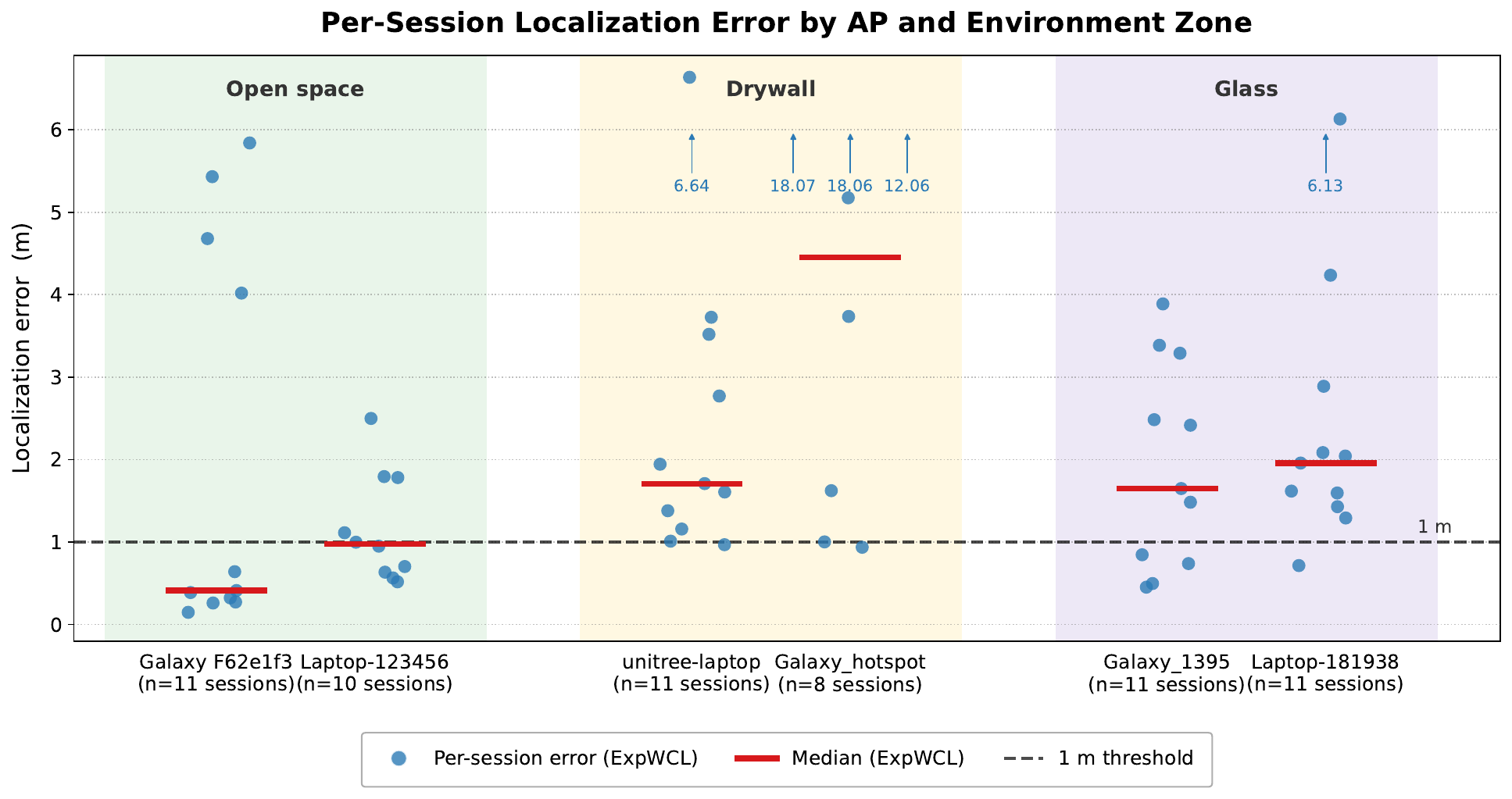}
  \caption{Per-session ExpWCL error for each AP, grouped by propagation environment. Each dot is one patrol; red bars mark per-AP medians. APs with multi-sided waypoint coverage show lower, more stable error than those primarily observed from one direction.}
  \label{fig:zone_strip}
\end{figure}

The primary operational question is the accuracy achievable from a single autonomous patrol without prior RF knowledge.
Table~\ref{tab:results} summarizes aggregate performance across 62 AP–session pairs.
Figure~\ref{fig:zone_strip} visualizes per-session ExpWCL error grouped by propagation environment.

\begin{table}[t]
\centering
\caption{Single-patrol localization performance across 62~AP--patrol pairs. ExpWCL achieves the lowest median and highest sub-1\,m rate. The 0.33\,m gap between ExpWCL and WCL indicates that estimator choice is a secondary factor. Filt.~LS+Huber degrades under the zero-prior constraint.}
\label{tab:results}
\setlength{\tabcolsep}{5pt}
\begin{tabular}{lrrrr}
\toprule
Method          & Median\,(m) & Mean\,(m) & Std\,(m) & \%\,$<$1\,m \\
\midrule
ExpWCL          & \textbf{1.62} & 2.71 & 3.49 & \textbf{32.3\%} \\
Ensemble        & 1.77 & 2.67 & 3.30 & 27.4\% \\
WCL             & 1.95 & 2.75 & 3.21 & 21.0\% \\
Filt.\ LS+Huber & 3.41 & 3.99 & 4.05 & 19.4\% \\
\bottomrule
\end{tabular}
\end{table}

ExpWCL achieves a median localization error of 1.62\,m with 32.3\% of estimates within 1\,m.
At this accuracy, the system reliably identifies the correct room or corridor segment of the rogue AP, which is sufficient to direct physical investigation without manual RF sweeping.

Ensemble and WCL follow at 1.77\,m and 1.95\,m respectively, a gap of 0.33\,m between the most complex centroid combination and the zero-parameter baseline.
This small gap indicates that estimator choice has a limited impact relative to measurement geometry.

Filt.~LS+Huber reaches a median error of 3.41\,m under the zero-prior constraint.
Without prior calibration, jointly estimating transmit power, path-loss exponent, and position increases solution variance and reduces stability relative to centroid methods.

\noindent\textbf{Patrol geometry as the primary accuracy determinant.}
As illustrated in Figure~\ref{fig:zone_strip}, APs with multi-sided waypoint coverage exhibit lower and more stable localization error than those observed primarily from one direction.
Open-space APs achieve medians of 0.41\,m and 0.97\,m, while glass-partition APs at the route periphery sit at 1.65\,m and 1.96\,m.
Among drywall APs, unitree-laptop (1.71\,m) benefits from central coverage, while Galaxy~hotspot (4.46\,m) represents a worst-case scenario.

The elevated error for Galaxy~hotspot arises from two factors.
First, MAC address rotation fragments observations across sessions, reducing the effective sample size.
Second, the enclosed position produces an RSSI field whose apparent peak is displaced toward the adjacent corridor along the dominant propagation path.
This highlights a limitation of RSSI-based localization under identifier instability and multipath.

No systematic drift is observed across time-of-day patrols, indicating temporal stability.
From an operational perspective, waypoint placement around high-risk zones is more impactful than selecting an algorithm.

\subsection{Multi-Patrol Aggregate}
\label{subsec:aggregate}

When multiple patrols are available, measurements can be fused to reduce variance.
We retain sessions whose WCL estimates are mutually consistent (within 1.9\,m) and construct an aggregate dataset using maximum RSSI per waypoint.

Five of six APs achieve sub-1\,m accuracy: open-space APs reach 0.29\,m and 0.45\,m; drywall APs reach 0.61\,m and 0.94\,m; one glass AP reaches 0.98\,m.
Galaxy\_1395 remains at 1.43\,m due to limited spatial coverage.

These results show that localization improves with additional consistent measurements when spatial diversity is sufficient.
Single-patrol performance remains the operational baseline; aggregation is an optional enhancement.

\begin{figure*}[t]
  \centering
  \includegraphics[width=\textwidth]{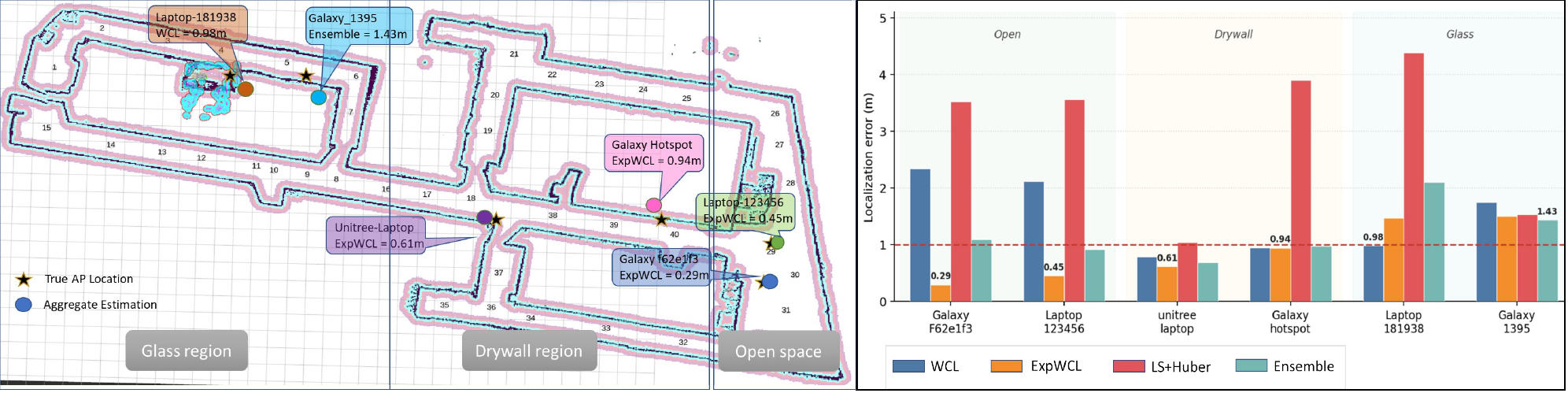}
  \caption{Aggregate localization results. Five of six APs fall below 1\,m under the best estimator.}
  \label{fig:aggregate}
\end{figure*}

\subsection{Blind Validation}
\label{subsec:blind}

We evaluate end-to-end generalization in a blind trial in which two APs are repositioned to unknown locations.
A single patrol is conducted without prior knowledge.
The system observes 73 BSSIDs and flags 8 (11.0\%) as potential rogue devices.
Both planted APs are correctly identified (100\% recall), with 6 additional candidates requiring verification.
Localization errors are 0.34\,m and 1.84\,m, consistent with single-patrol performance.
Six additional locally-administered BSSIDs are flagged.
This corresponds to a manageable false-positive set in practice, as flagged devices can be quickly verified by physical inspection once localized.

\begin{table}[t]
\centering
\caption{Blind validation results.}
\label{tab:blind}
\small
\begin{tabular}{lccrr}
\toprule
AP & Position & Det. & Method & Err.\,(m) \\
\midrule
Galaxy F62e1f3 & (17.4, $-$11.5) & \ding{51} & Ensemble & 0.34 \\
Galaxy\_1395   & ($-$1.0, $-$4.0) & \ding{51} & WCL & 1.84 \\
\bottomrule
\end{tabular}
\end{table}

\subsection{Robustness to Patrol Incompleteness}

We evaluate sensitivity to waypoint coverage by subsampling the patrol route.
Median error increases from 1.62\,m to 2.02\,m at 64\% coverage and to 3.07\,m at sparse sampling.

Performance remains stable down to moderate coverage levels, after which degradation accelerates, indicating a minimum spatial density requirement.

\begin{figure}[t]
\centering
\includegraphics[width=\linewidth]{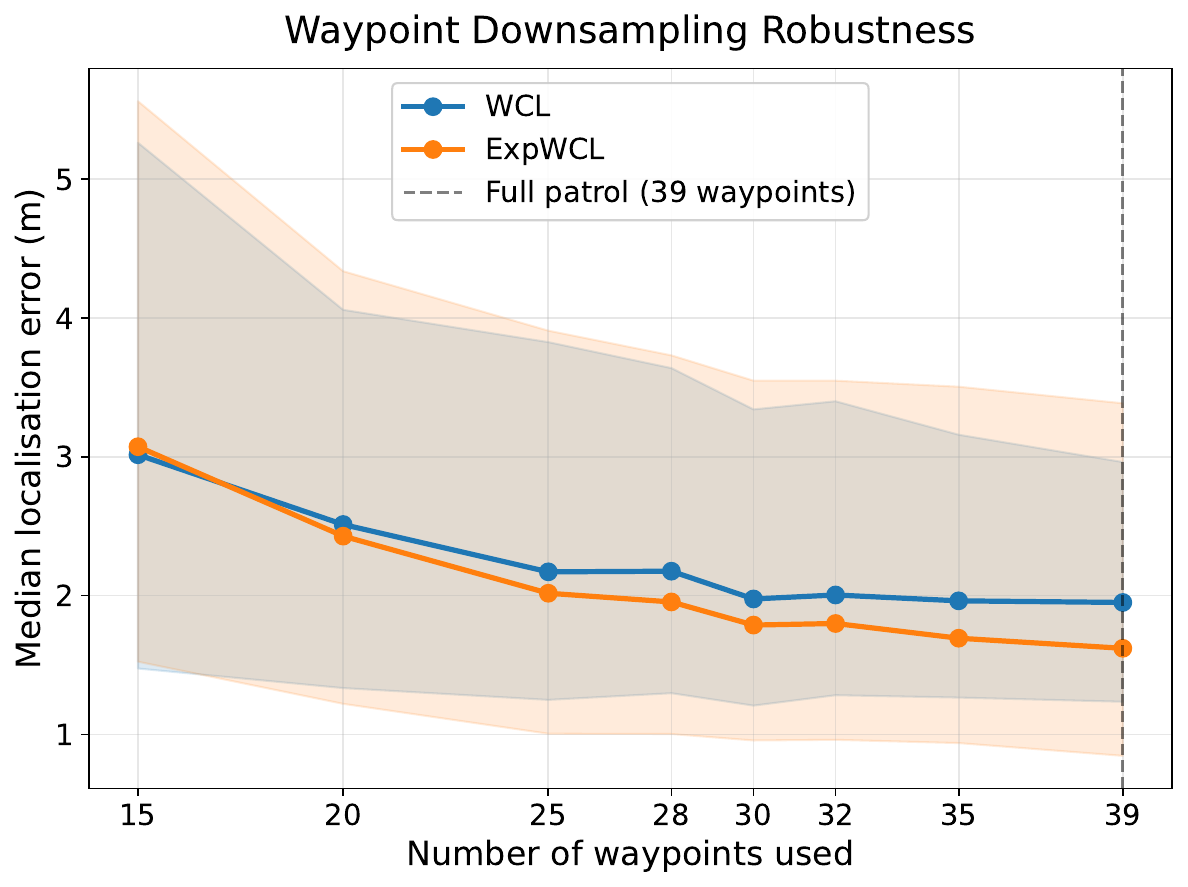}
\caption{Localization error vs. patrol coverage.}
\label{fig:waypoint_robustness}
\end{figure}

\subsection{Limitations}
\label{subsec:limits}

The evaluation is limited to a single building over four days.
All APs are stationary and non-adaptive.
The system does not address MAC spoofing or adversarial evasion.
Performance under multi-floor environments, dense interference, or dynamic attackers remains future work.
\section{Discussion}
\label{sec:discussion}

\noindent\textbf{Feasibility.}
The evaluation demonstrates that autonomous, zero-configuration localization of rogue wireless devices is feasible under realistic operational constraints.
A single 20-minute patrol localizes detected devices to a median of 1.62\,m, which is sufficient to identify the correct room or corridor segment for targeted physical inspection.
In the blind trial, previously unseen placements are localized to 0.34\,m and 1.84\,m, consistent with the primary results.
These findings indicate that infrastructure-free localization can support rapid physical incident response without requiring pre-installed sensing systems or RF calibration.

\noindent\textbf{Estimator complexity.}
Across 62 single-patrol sessions, ExpWCL improves upon the zero-parameter WCL baseline by 0.33\,m in median error.
All centroid-based estimators operate on the same RSSI geometry and therefore exhibit similar performance.
In contrast, parametric path-loss fitting reaches a median error of 3.41\,m under the zero-prior constraint.
Without calibration, joint estimation of transmit power, path-loss exponent, and position increases solution variance.
This suggests that under incident response conditions, where prior information is unavailable, lightweight centroid estimators provide a more stable and reliable solution than model-based approaches.

\noindent\textbf{Patrol geometry.}
Localization error is primarily governed by spatial measurement coverage rather than obstruction category.
APs surrounded by waypoints from multiple directions achieve sub-meter accuracy, whereas APs near patrol peripheries exhibit larger errors even with repeated measurements.
Multi-patrol aggregation reduces variance when coverage is sufficient, but single-patrol performance remains the operational baseline.
From a deployment perspective, this implies that patrol design—specifically, waypoint placement around high-risk zones—has greater impact than estimator selection.

\noindent\textbf{Operational considerations.}
The system complements existing wireless intrusion detection mechanisms by enabling physical localization after detection, allowing security teams to transition from alert generation to actionable remediation.
Detection produces a small candidate set that can be verified by physical inspection once localized.
The full pipeline—from detection to localization—can be completed within a single patrol cycle, enabling practical integration into incident response workflows in enterprise or industrial environments.

\noindent\textbf{Limitations.}
The current evaluation assumes stationary, non-adaptive devices emitting standard beacon frames.
Adversaries that disable transmission, rotate identifiers, or actively evade detection may reduce the continuity of the observable signal and degrade localization performance.
Additionally, the study is limited to a single-floor environment and does not explicitly model multi-floor interference or highly dense wireless environments.
These factors represent important directions for future work.

\section{Conclusion}
\label{sec:conclusion}

We presented \sys{}, an autonomous system for localizing rogue wireless devices using a single commodity mobile robot.
The system operates without RF fingerprinting, pre-installed sensors, or site-specific calibration, enabling deployment in environments where prior infrastructure is unavailable.

In an 800\,m$^2$ indoor environment, a single 20-minute patrol achieves a median localization error of 1.62\,m using passive beacon measurements and centroid-based estimators.
A blind evaluation confirms generalization to unseen device placements, correctly identifying rogue devices among 73 observed BSSIDs and localizing them with errors of 0.34\,m and 1.84\,m.
With multi-patrol aggregation, five of six devices are localized within 1\,m.

The key observation is that spatial measurement coverage, enabled by autonomous patrol, has a greater influence on localization accuracy than estimator complexity.
Under zero-prior conditions, simple centroid-based methods provide robust and practical performance without requiring calibration or model tuning.

These results demonstrate that infrastructure-free localization is a viable approach for enabling physical incident response in cyber-physical environments, bridging the gap between network-level detection and physical remediation.

\bibliographystyle{splncs04}
\bibliography{references}

\end{document}